\documentclass{article}
\usepackage{ijcai22}

\usepackage{times}

\usepackage{soul}
\usepackage{xcolor}
\usepackage{url}
\usepackage[colorlinks=true,linkcolor=blue]{hyperref}
\usepackage[utf8]{inputenc}
\usepackage[small]{caption}
\usepackage{graphicx}
\usepackage{amsmath}
\usepackage{booktabs}
\usepackage{adjustbox}
\usepackage{listings}
\usepackage{soul}
\usepackage{amssymb}% http://ctan.org/pkg/amssymb
\usepackage{pifont}% http://ctan.org/pkg/pifont
\newcommand{\cmark}{\ding{51}}%
\newcommand{\xmark}{\ding{55}}%
\title{A Mental-Model Centric Landscape of Human-AI Symbiosis}
\author{
Zahra Zahedi\and
Sarath Sreedharan\And
Subbarao Kambhampati
\affiliations
School of Computing and AI, ASU\\
\emails
\{zzahedi, ssreedh3, rao\}@asu.edu
}
\begin{document}

\maketitle

\begin{abstract}
There has been significant recent interest in developing AI agents capable of effectively interacting and teaming with humans. 
While each of these works try to tackle a problem quite central to the problem of human-AI interaction, they tend to rely on myopic formulations that obscure the possible inter-relatedness and complementarity of many of these works.
The human-aware AI framework was 
a recent effort to provide a unified account for human-AI interaction by casting them in terms of their relationship to various mental models.
Unfortunately, the current accounts of human-aware AI are insufficient to explain the landscape of the work doing in the space of human-AI interaction due to their focus on limited settings.
In this paper, we aim to correct this shortcoming by introducing a significantly general version of human-aware AI interaction scheme, called {\em generalized human-aware interaction (GHAI),} that talks about (mental) models of six types. Through this paper, we will see how this new framework allows us to capture the various works done in the space of human-AI interaction and identify the fundamental behavioral patterns supported by these works. 
We will also use this framework to identify potential gaps in the current literature and suggest future research directions to address these shortcomings.
\end{abstract}
\section{Introduction}
As AI systems are becoming an increasing part of our day-to-day lives, there is a growing interest in designing AI systems that can interact with and team with humans fluently and effectively.
As such, there is an increasing body of scholarly works that have tried to address various challenges related to human-AI interaction in different settings.
The richness and diversity of challenges related to human-AI interaction mean the works tend to use diverse tools and formulations to address the challenges they are trying to address.
Unfortunately, this has also meant it is hard to make sense of the overall landscape of the field owing to the lack of a unifying framework, a shortcoming we hope to address in this paper.
%. We hope to address this lack of a unifying framework in this paper.

Within psychology, mental models and in particular the theory of mind \cite{premack1978does} have been proposed as a mechanism to make sense of human-human interaction.
One could potentially extend the idea theory of mind and the use of mental models to also ground and categorize interaction between humans and AI systems.
The framework of human-aware AI \cite{sreedharan2022explainable} has been proposed to incorporate the intuition of theory of mind into the context of human-AI interaction. Unfortunately, the current versions of the human-aware AI framework are insufficient to adequately explain the various works in the area. 
%This is because the original human-aware AI framework was introduced to capture interactions possible in scenarios where you have the human limited to the role of a passive observer trying to make sense of an AI agent's decisions. 
This is in part because the original human-aware AI framework was introduced to capture settings where the role of the human is limited to being a passive observer trying to make sense of an AI agent's decisions. 
These scenarios only represent a small subset of possible ones that have been considered by the various works studied within the purview of human-AI interaction.
\begin{figure}[t]
    \centering
    \includegraphics[width=.5\textwidth]{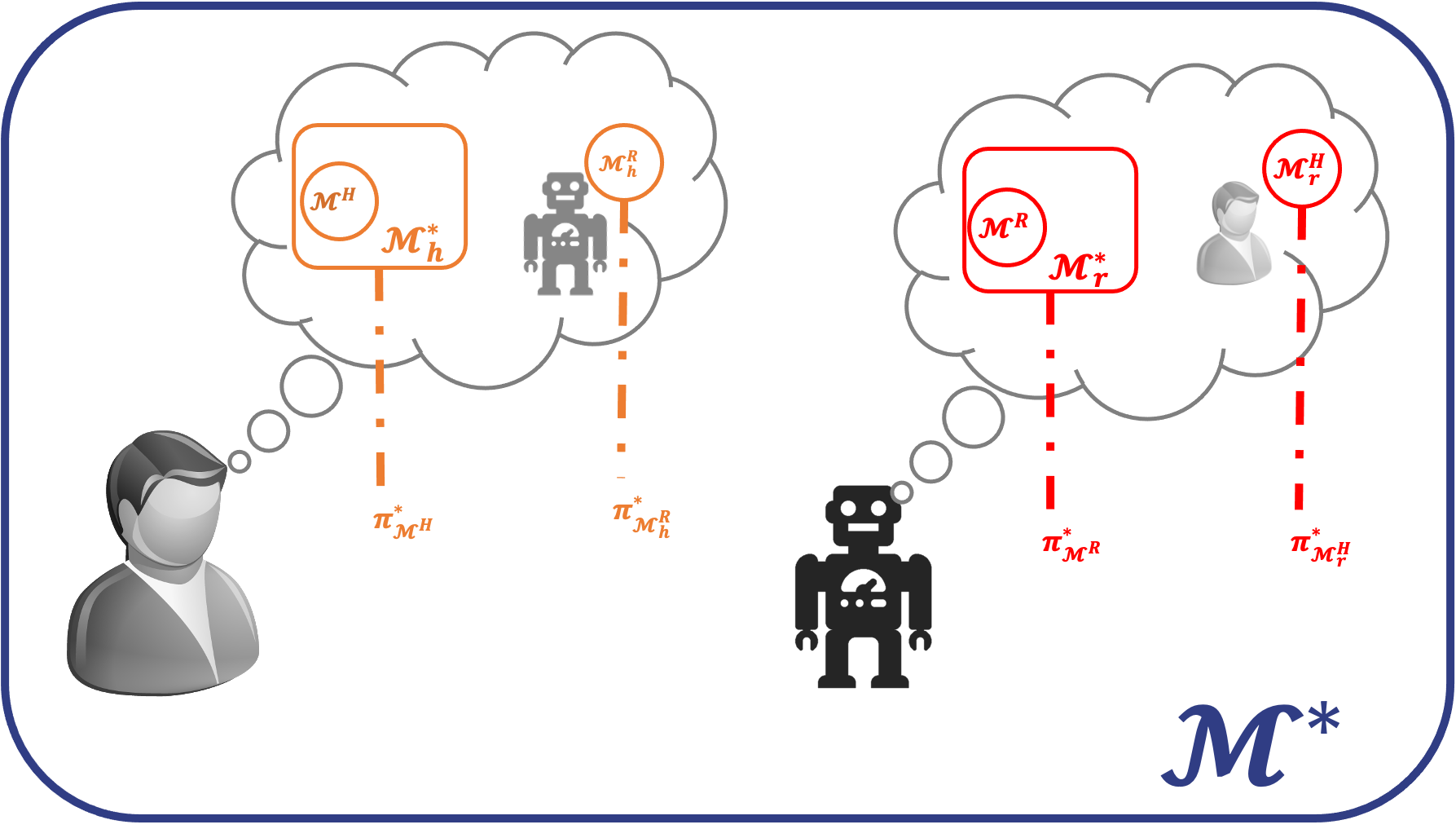}

\caption{The six models in the GHAI framework. $\mathcal{M}^*$ are the ground truth models of the task; $\mathcal{M}^H$ and $\mathcal{M}^R$ are the task models that the human and the AI agent ascribe to themselves; $\mathcal{M}^R_h$ and $\mathcal{M}^H_r$ are the estimates of the AI agent's (human's) model that human (AI agent) has.}
    %\caption{A diagrammatic representation of the Generalized Human-Aware interaction framework. GHAI captures the interactions between a human and the AI agent. Right side shows the models held by the AI agent and left those by the human.}

    \label{fig1}
%   \vspace{-10pt}
\end{figure}

In this paper, we present a more general version of the human-aware AI interaction framework (Figure \ref{fig1}). GHAI will not only allow for scenarios where the human may be an actor but also introduce the notion of a task model $\mathcal{M}^*$ that captures the true joint task specification of both the human and AI agent. Moreover, we will consider each agent's (i.e. human or the AI agent) perception of the true combined task, which is independent of their beliefs about the other agent's perception of it. This allows us to capture interaction scenarios where one of the agents may choose to correct the other agent's beliefs about their task models. With the basic framework in place, we will see how the primary interaction facilitated by the various works is manifestation of either a {\em model-communication behavior} or a {\em model-following behavior} made possible by the use of the various mental models that are part of the framework.

This paper brings together works from various communities that are studied under different problem frameworks including human-aware planning, human-AI cohabitation, human-robot interaction, human-in-the-loop ML, and human-AI symbiosis. The works considered here look at facilitating a diverse set of interactions, including explanation, teaming, teaching, and debugging, and have been presented in forums like  CHI, HRI, ICAPS, IJCAI, ICML, and HCOMP. By bringing together this rich and varied body of works under a single umbrella, we highlight the centrality of the role played by mental models in the context of human-AI interaction and suggest direction.

%%%%%%%%%%%%%%%%%%%%%%%%%%%%%%%%%%%%%%%%%%%%%%%%%%%%%%%%%%%%%%%%%%%%%%%%%%
% The rest of the paper is structured as follows. Section \ref{GHAI} will introduce the generalized human-aware AI interaction framework and describe the various task models that are part of the framework. In section \ref{landscape}, we will use the framework to characterize the primary interactions facilitated by the various prominent works in the area of human-AI interaction. Finally, we will conclude the paper with Section \ref{discussion} which will present an extended discussion on some of the overlooked problems in the area, made more prominent by the introduction of the generalized framework, and also discuss some related challenges that need to be addressed to create truly human-aware systems.
%%%%%%%%%%%%%%%%%%%%%%%%%%%%%%%%%%%%%%%%%%%%%%%%%%%%%%%%%%%%%%%%%%%%%%%%%
% \textbf{Survey Scope and Outline} In this survey, we propose a generalized human-aware interaction framework that captures all necessary models of each agent for their interaction, and then we provide a taxonomy related to the introduced frameworks. They includes model information from different models and model following behavior. So, we highlight how different works in the area of human-AI interaction can be fit and organized in these dimensions. Furthermore, given the introduced framework, we discuss the emerging gaps in the current literature that can be useful for further research in this area.

\section{Generalized Human-Aware Interaction}
\label{GHAI}
In this section, we describe the basic interaction framework that we will be leveraging to analyze the works that we are considering in this paper. The {\em Generalized Human-Aware Interaction} builds on earlier work done within the context of human-aware planning \cite{sreedharan2022explainable}, that was introduced to characterize interpretable behavior generated in the context of a planning agent trying to generate behavior in the presence of a human observer. The framework allowed the agent to leverage human's expectations about the agent in its decision-making process, by capturing these expectations in the form of a mental model that the human could be thought to ascribe to the agent. 
%Unfortunately, when we take a look at the wider landscape of works done to model and understand interaction between humans and system, the existing human-aware planning framework falls short in capturing them. 
Unfortunately, the existing human-aware planning framework is insufficient to explain the wider landscape of works done in the area of human-AI interaction.
So in this paper, we introduce {\em Generalized Human-Aware Interaction (GHAI)} that significantly expands the framework in multiple dimensions to explain and capture the extant literature.
Figure \ref{fig1} presents a diagrammatic overview of GHAI and in the rest of the section, we will provide an overview of the various components of the system.

\indent\textbf{Agents--} 
Our central modeling objective is to capture the interaction between a human denoted by $H$ and AI agent denoted by $R$ (standing for robot, which we use in a more general sense). While our framework can be extended to cases when we have multiple humans and multiple AI agents, we will focus on this dyadic interaction to allow for more concise exposition. Unlike previous frameworks, we place no restrictions on either the roles played by the agents in the context of the interaction, nor any restrictions on how the objectives of each agent may or may not be related.

\textbf{Model--} 
The central tool we will be using to capture both the agent decisions and agent's expectations from others, will be a task model.
A task model (usually denoted as $\mathcal{M}^{\cdot}_{\cdot}$) is any mathematical model that encodes among other things, an entities beliefs about task objectives, state of the world and how the world may evolve on its own or in response to an agent action. There exist a wide variety of options for what the models might be (for example it may be MDPs \cite{puterman2014markov}, differential system equations or symbolic models like PDDL \cite{geffner2013concise}), but for the purposes of this paper, their exact form is irrelevant. We only require that the models used by an agent is in a form that they can use to derive the required decision or expected decisions. % i.e., it is in a form that the agent can reason with it and it captures the required information. 
%%%%%%%%%%%%%%%%%%%%%%%%%%%%%%%%%%%%%%%%%%%%%%%
%cite for mdp  \cite{puterman2014markov} also cite pddl
%%%%%%%%%%%%%%%%%%%%%%%%%%%%%%%%%%%%%%%%%%%%%%%%%
As apparent from Figure \ref{fig1}, GHAI consists of interaction between multiple models. 
For each individual model, we consider that we have to talk about the entity the model is ascribed to (i.e., the entity that is being described by the model) and the entity that the model belongs to (i.e., the entity that is using the model). In particular, we are interested in three entities, the two agents (denoted as $H$ and $R$) and the true ground truth task (denoted with $*$). We will be using the superscript to denote the entity described by the model and the subscript will describe the entity that the model belongs to. As additional naming convention, we will use lowercase in the case of subscripts and drop it completely when both superscript and subscript matches. In particular, we only need to consider the following models.

\textbf{Task models $\mathcal{M}^*_{\cdot}$--} This model is meant to capture the entirety of the task. As such it consists of all of the actions, objectives, preferences of both agents and additional facts about the world state that may not fall into the purview of the individual agent models. In this class, we have three specific models to consider. The ground truth model $\mathcal{M}^*$ ( same as saying $\mathcal{M}^*_*$) that is unknown to either of the agents, that captures the true objective and dynamics of the task. This model could be thought of as belonging to nature or to the true task-specifier that is not directly engaging in the interaction. Next we have the task model maintained by each of the agents ($\mathcal{M}^*_r$ and $\mathcal{M}^*_h$). These models can be thought of as the union of their beliefs about their own model (will be covered later) and their beliefs about what the true model of the other agent should be and any additional task facts. 
%One could see that when each agent is making a decision they are in fact considering this estimate of the task model 
%facts about the world that is true but might not be known to the agents.

\textbf{Model of each agent $\mathcal{M}^R$ and $\mathcal{M}^H$--} This is the model each agent ascribes to themselves. This determines what actions each agent believes they could perform and the objectives and preferences they are trying to satisfy. These models are also part of each agent's beliefs about what the true task model is, i.e., $\mathcal{M}^R$ is part of $\mathcal{M}^*_r$ and $\mathcal{M}^H$ is part of $\mathcal{M}^*_h$.

\textbf{One agent's model of the other $\mathcal{M}^R_h$ and $\mathcal{M}^H_r$--} These consist of what one agent believes the model of the other agent is, i.e, $\mathcal{M}^R_h$ is the human's belief about $\mathcal{M}^R$, and $\mathcal{M}^H_r$ is the AI agent's belief about $\mathcal{M}^H$.
Note that we don't expect these models to be same as what each agent thinks the true capabilities and objectives of the other agent may be. For example, from the perspective of the AI agent, what the AI agent thinks the human capabilities and objective should be (as encoded in $\mathcal{M}^*_r$) need not to be the same as what the AI agent thinks the human ascribes to themselves. These models capture the theories of mind captured by each agent.

These are all models we will need to consider to describe the space of works and various challenges related to human-AI symbiosis. One could of course consider additional models by including more nested beliefs,
%(for example, human's beliefs about robot's beliefs about the human), 
but in most practical scenarios people generally do not consider too many levels of nesting in their reasoning \cite{wright2017predicting} and as we will see from our survey most existing works tend to avoid such nesting too. Next we will move onto the various decisions that may be derived from these various models. 

\textbf{Decisions} 
%These correspond to various decisions that are derived in accordance with the various model. 
We will generally denote decisions as $\pi_{\cdot}$ with the subscript corresponding to the model from which it is derived. Again, we will not limit ourselves to a specific decision type and depending on the exact problem the space of decisions could vary from single shot labels one may associate with a classification task to a policy to be carried out by an embodied agent. In some cases, there might also be multiple possible decisions that may be allowed by the model (referred to as valid decisions) and in these cases the model may also enforce an ordering among the possible decisions. We will refer to a decision as being optimal for a given model if there exist no other decision that may be preferred over the given one. The various decisions, we will consider in GHAI setting include,

\textbf{Decisions made by the actor $\pi_{\mathcal{M}^{R/H}}$--} These are the decisions made by each agent to be carried out in the world (so these  come into play when the agent is an actor). These decisions are primarily generated by using $\mathcal{M}^{R/H}$ but depending on the scenario they may also consider their belief about the other agent's task model.
%about the true task model $\mathcal{M}^*_{r/h}$ or the model they believe the other agent ascribes to itself $\mathcal{M}^{H/R}_{r/h}$  

\textbf{The agent belief about the decision made by the other agent actor $\pi_{\mathcal{M}^{R/H}_{h/r}}$--} These are the decisions that an agent might ascribe to the other agent that is acting. They are primarily derived using their beliefs about the other agent ($\mathcal{M}^R_h$ or $\mathcal{M}^H_r$).\looseness=-1

\section{Landscape of Human-AI Symbiosis}
\label{landscape}
The models described in the previous section could come into play in human-AI interaction in various roles. In particular, one can see the models being used in two critical ways, namely, to drive a {\em model-communication} behavior wherein one of the agent uses it's beliefs about the task to mould the belief of the other agent or to drive a {\em model-following} behavior wherein the agent chooses behaviors that considers the beliefs of the other agent. 
While previous works in human-aware AI have noted limited forms of such behavior \cite{sreedharan2020unify}, our focus on the larger landscape of human-AI interaction allows us to view these behavioral patterns in a much more general way.
%by considering the full breadth of the works performed in the context of Human-AI symbiosis, we are able to better appreciate the various unique ways these general behavioral trends could manifest in the context of human-AI interaction.
Thus we will try to categorize and analyze the works done in this space along these two dimensions. In the case of {\em model-communication} we will further break the works down based on the source of communication, i.e., is the AI-system trying to communicate to the human (thus the information originates from $\mathcal{M}^*_r$) or is the human trying to communicate to the AI-system (thus the information originates from $\mathcal{M}^*_h$). In the case of communication we will consider both explicit communication where the agents are directly providing model information or may engage in implicit communication where the agent tries to influence the other agent's beliefs through their choice of behavior. Note that these categories aren't mutually exclusive and as we will see many works in this space support behavior from multiple categories.

\subsection{Model Information from AI Agent's End ($\mathcal{M}^*_r$)} 
In this section, we have works that mostly use the models from the right side of Figure \ref{fig1} (AI mental models) to update models held by the human. For this we start with model information from $\mathcal{M}^*_r$ that includes both their own AI model and their beliefs of the other agent true model. 
Within this category, we can further categorize works based on the human models that will be updated (i.e., $\mathcal{M}^R_h$ or $\mathcal{M}^H$) and whether or not the AI agent makes use of a mental model of the human ($\mathcal{M}^H_r$) to generate the required information (see the summary in Table \ref{tab1}).

\subsubsection{Updating $\mathcal{M}^R_h$}
The human uses $\mathcal{M}^R_h$ to make sense of the AI's decisions. So if the AI's model is different from what the human believes of them, this can result in the human finding the AI behavior incomprehensible or surprising \cite{chakraborti2019explicability}. For this reason, when the AI agent needs to ensure that the human is able to correctly evaluate AI decisions, it could use information from $\mathcal{M}^*_r$ to update  $\mathcal{M}^R_h$, either implicitly or explicitly.

Examples of implicit model communication, include legible behavior generation or transparent planning which enables the human collaborators to infer the goal purely from the agent behavior \cite{dragan2013legibility,dragan2013generating,dragan2015effects,macnally2018action}. Implicit communication can also be used to obfuscate the plans and objectives of the AI agents to protect its privacy from adversarial entities \cite{kulkarni2019unified} or to synthesize behaviors that are simultaneously legible to friendly entities and obfuscatory to adversarial ones \cite{kulkarni2019signaling}. The AI agent can also update $\mathcal{M}^R_h$ to inform the human about its limitations, by implicitly communicating its incapabilities through showing how and why it is unable to accomplish its current objectives \cite{kwon2018expressing}. It can also use non-verbal cues and feedback to signal how it expects the human to act next to communicate their preferences more effectively \cite{huang2019nonverbal}.

On the other hand, one of the common cases where explicit communication is used to update $\mathcal{M}^R_h$ is in the context of explanations.
The explanation may involve analyzing and explaining the details of failures and errors \cite{nushi2018towards}. Such model updates can be good for debugging and troubleshooting purposes. 
The explanation can also solve the root cause of inexplicable behavior, by reconciling the difference between $\mathcal{M}^R_h$ and AI model $\mathcal{M}^R$ until the behavior becomes explicable to the human
\cite{chakraborti2017plan,sreedharan2017explanations,sreedharan2018hierarchical,zahedi2019towards,soni2021not}, 
%%%%%%%%%%%%%%%%%%%%%%%
%cite utkarsh here ,soni2021not
%%%%%%%%%%%%%%%%%%%%%%%%%%%%
%or update the model up to a point to balance between the updated model and generated behavior 
or update $\mathcal{M}^R_h$ until it can support aligned behavior (to be covered in detail later)
\cite{chakraborti2017balancing,sreedharan2019expectation}. Also, if the behavior expected by the human is infeasible, the explanation updates the model by expressing incapability. This can be done by presenting unsolvability cores for specific behaviors \cite{raman2013sorry,raman2013towards,briggs2015sorry,sreedharan2019can}. In the case of a joint workspace where both the human and robot acts, communicating the policy can update the human model of AI agent $\mathcal{M}^R_h$ for better cooperation \cite{unhelkar2020decision}.

\subsubsection{Updating $\mathcal{M}^H$}
In this case human is an actor and the robot is trying update the human's model with what they believe to be information about the true underlying task. This update can again happen implicitly or explicitly. An example of implicit model update is when the robot makes a serendipitous plan to make the task easier for the human \cite{chakraborti2015planning}. Explicit model updates happen in cases like decision support systems such as RADAR where the human-decision maker may be provided the previously unknown task information to aid their decision-making process of single
%that improve the decision-making experience of the human by providing suggestions that aid in constructing a plan for single 
\cite{sengupta2017radar,grover2020radar,valmeekam2020radar} and multiple \cite{sengupta2018ma} humans. 
It is also possible that the AI agent may be trying to update both $\mathcal{M}^R_h$ and $\mathcal{M}^H$. 
%This might be the case for joint work space condition with human and AI agent that both are acting. When the AI agent communicate their intentions and actions to update $\mathcal{M}^R_h$ 
This might be the case in joint workspaces where the AI agent needs to both communicate its intentions and actions to update $\mathcal{M}^R_h$ 
and express what it expects the human to do and thus update $\mathcal{M}^H$ \cite{gao2020joint}.

\subsubsection{Use of $\mathcal{M}^H_r$}
Here we will look at some works that allow the AI agent to use it's mental model of the human when selecting the information to be provided. This includes both cases where the AI system is trying to update $\mathcal{M}^R_h$ or $\mathcal{M}^H$.
The former corresponds to cases when the behavior (implicit update) or explanation (explicit update) is personalized to the human model and preferences. Examples include providing faithful and customized explanations of a black box classifier that accounts for the fidelity to the original model as well as user interest \cite{lakkaraju2019faithful}, 
%or personalized explanations to the user for sequential decision making tasks performed by a robot in which there might be different user types 
explanations expressed in the human's vocabulary
\cite{sreedharan2020bridging}, or choosing explanation based on the human's preference to the type of information \cite{zahedi2019towards}.
%different preferences over types of explanation \cite{zahedi2019towards}.

All works discussed above that allow for the update of $\mathcal{M}^H$ use $\mathcal{M}^H_r$ as basis for the AI agent to decide what information to provide to the human. However it is worth noting that all the works discussed here implicitly assume that the AI agent's belief about the human model is the same as the human's own model, i.e.,  $\mathcal{M}^H_r$ is same as $\mathcal{M}^H$.
% When the human is the main actor and making the final decision, the AI agent can use the model information from $\mathcal{M}^*_r$ to help human in their task. This help will update human model of the task $\mathcal{M}^H$ while the AI model of the human is used $\mathcal{M}^H_r$ because the AI agent has directly access to this instead of $\mathcal{M}^H$. This update can again happen implicitly or explicitly. An example of implicit model update is when the robot make a serendipitous plan to make the task easier for the human \cite{chakraborti2015planning}. Explicit model updates happen in cases such as proactive decision-making systems that improves the decision-making experience of the human by providing suggestions that aid in constructing a plan for single \cite{sengupta2017radar,grover2020radar,valmeekam2020radar} and multiple \cite{sengupta2018ma} humans. 

There might also be cases where the AI system may be trying to update both $\mathcal{M}^H$ and $\mathcal{M}^R_h$ while Using $\mathcal{M}^H_r$. One example is the AI task allocation system AITA  \cite{zahedi2020not} that uses a negotiation based algorithm to fairly allocate different tasks to humans and giving contrastive explanation (using $\mathcal{M}^H_r$) to update both  $\mathcal{M}^H$ and $\mathcal{M}^R_h$, where $\mathcal{M}^R_h$ captures team performance and constraints.
\begin{table}[t]
\centering
\caption{Summary of the various aspects that come into play in cases where model information is coming from $\mathcal{M}^*_r$
% \textbf{\textit{(Note for reviewers: We used the plain bib format so we can tabulate the paper numbers)}}
}
\label{tab1}
\begin{adjustbox}{width=0.5\textwidth}

\begin{tabular}{p{30mm}|c|c|c}
\toprule
References & \multicolumn{3}{c}{Model information from $\mathcal{M}^*_r$}\\
\cmidrule{2-4}
&$\mathcal{M}^H_r$ Used & Updating $\mathcal{M}^H$& Updating  $\mathcal{M}^R_h$\\
\cmidrule{1-4}
\cite{dragan2013legibility,dragan2013generating,dragan2015effects,macnally2018action,kulkarni2019unified,kulkarni2019signaling,kwon2018expressing,huang2019nonverbal,nushi2018towards,chakraborti2017plan,sreedharan2017explanations,sreedharan2018hierarchical,chakraborti2017balancing,sreedharan2019expectation,raman2013sorry,raman2013towards,briggs2015sorry,sreedharan2019can,unhelkar2020decision,soni2021not}&\xmark & \xmark& \cmark\\
\cmidrule{1-4}
\cite{lakkaraju2019faithful,sreedharan2022explainable,zahedi2019towards}&\cmark & \xmark& \cmark\\
\cmidrule{1-4}
\cite{gao2020joint}&\xmark & \cmark& \cmark\\
\cmidrule{1-4}
\cite{chakraborti2015planning,sengupta2017radar,grover2020radar,valmeekam2020radar,sengupta2018ma}&\cmark & \cmark& \xmark\\
\cmidrule{1-4}
\cite{zahedi2020not}&\cmark & \cmark& \cmark\\
\bottomrule
\end{tabular}

\end{adjustbox}
\vspace{-0.2cm}
\end{table}
\subsection{Model Information from Human's End ($\mathcal{M}^*_h$)} 
In this section, we will consider works consider the use of the left side of Figure \ref{fig1} (Human mental models) to update AI agent models. Thus, we have model information coming from $\mathcal{M}^*_h$ that is used to update $\mathcal{M}^R$, $\mathcal{M}^H_r$, or both, and the human might use $\mathcal{M}^R_h$ to select this information (Table \ref{tab2}). 
\subsubsection{Updating $\mathcal{M}^R$}
First we will consider works that use model information and inputs from human models to build, update and improve AI agent model $\mathcal{M}^R$. The challenge tackled by works that allow for such model updates is not only to facilitate the update of $\mathcal{M}^R$ but also in deciding how and when to use the model information from $\mathcal{M}^*_h$ \cite{kamar2016directions}. 
% Basically, the model information from $\mathcal{M}^*_h$ can be used to update and improve AI agent model $\mathcal{M}^R$ accuracy in solving a task. 
One common example is the problem of crowdsourcing task information from a group of people to create a more accurate task model for a single shot machine learning task (e.g. systems like Crowdsynth and CrowdExplorer) \cite{kamar2013light,kamar2013lifelong,kamar2015planning}, and the challenge here would be to identify the most effective way to collect the information to learn the task \cite{wilder2020learning,kamar2013light}. Other works have also looked at using similar settings to collect information about more complex tasks including planning \cite{gao2015acquiring}. Model information from, and decisions made using $\mathcal{M}^*_h$, could also be used indirectly to detect limitations and errors in the AI agent model $\mathcal{M}^R$, which can then be updated. Thus the information from $\mathcal{M}^*_h$ is used to troubleshoot and then update the model $\mathcal{M}^R$. For instance, works have looked at analyzing how human and machine decisions differ for the same problem and using that information to update the AI model. \cite{tan2018investigating} looked at using such an approach to improve the effectiveness of recidivism prediction, and \cite{nushi2017human} also looks at improving the AI systems performance by troubleshooting the systems failure, but performs this analysis and model update iteratively. Furthermore, the model information from $\mathcal{M}^*_h$ can be in the form of an advice or demonstration to the AI agent to update $\mathcal{M}^R$ \cite{amir2016interactive,torrey2013teaching,huang2019nonverbal,ramakrishnan2019overcoming}.\looseness=-1

\subsubsection{Updating $\mathcal{M}^H_r$}
The AI agent can also use the input and information from human $\mathcal{M}^*_h$ to make and refine the model they ascribe to the human ($\mathcal{M}^H_r$) \cite{ramakrishnan2019overcoming,talamadupula2017architectural}. Examples of this include cases where the AI agent uses the answers to its questions to refine its understanding of the human model $\mathcal{M}^H_r$ \cite{grover2020model}, or when the AI agent use the human critiques \cite{grover2020radar,sengupta2017radar,sengupta2018ma} and explanatory queries \cite{valmeekam2020radar} raised by the human to update and refine $\mathcal{M}^H_r$. 
\subsubsection{Use of $\mathcal{M}^R_h$}
Now, we will look at works that expects the human to use their mental model of the AI agent when providing the information. This includes both cases where the human information is used to update $\mathcal{M}^H_r$ or $\mathcal{M}^R$. The latter happens when the human is advising or teaching the AI agent while considering $\mathcal{M}^R_h$. The human can explicitly advise and give instruction to the AI agent by suggesting actions that the AI agent should take while learning \cite{amir2016interactive,torrey2013teaching}, or the human can use $\mathcal{M}^R_h$ to implicitly teach the AI agent through demonstration \cite{huang2019nonverbal}.\looseness=-1

There might be cases where the human information is used to update both $\mathcal{M}^R$ and $\mathcal{M}^H_r$ while using $\mathcal{M}^R_h$. Examples here include cases, where the human demonstration is used to detect blind spots of both the AI agent and the human. Here the demonstration is used by the AI agent to update $\mathcal{M}^R$ and $\mathcal{M}^H_r$, so it can hand off the task to the most capable agent \cite{ramakrishnan2019overcoming}. Another example includes the AI agent's use of human instructions to operate in open world scenarios \cite{talamadupula2017architectural}.
%There might be cases where the human information is used to update both $\mathcal{M}^R$ and $\mathcal{M}^H_r$ while using $\mathcal{M}^R_h$. Examples here include cases, where the human demonstration is used to detect blind spot of both the AI agents and the human to update $\mathcal{M}^R$ and $\mathcal{M}^H_r$ that enables handing off the task to the most capable agent \cite{ramakrishnan2019overcoming}, or when the AI agent is using human instructions to understand the different aspects of an unknown environment to handle acting in an open world \cite{talamadupula2017architectural}.
\begin{table}[t]
\centering
\caption{Summary of the various aspects that come into play in cases where model information is coming from $\mathcal{M}^*_h$}
\label{tab2}
\begin{adjustbox}{width=0.5\textwidth}
\begin{tabular}{p{30mm}|c|c|c}
\toprule
References & \multicolumn{3}{c}{Model information from $\mathcal{M}^*_h$}\\
\cmidrule{2-4}
&$\mathcal{M}^R_h$ Used & Updating $\mathcal{M}^R$& Updating  $\mathcal{M}^H_r$\\
\cmidrule{1-4}
\cite{grover2020model,grover2020radar,sengupta2017radar,sengupta2018ma,valmeekam2020radar}&\xmark & \xmark& \cmark\\
\cmidrule{1-4}
\cite{kamar2016directions,kamar2013light,kamar2013lifelong,kamar2015planning,kamar2013light,gao2015acquiring,tan2018investigating,nushi2017human}&\xmark & \cmark& \xmark\\
\cmidrule{1-4}
\cite{amir2016interactive,torrey2013teaching,huang2019nonverbal}&\cmark & \cmark& \xmark\\
\cmidrule{1-4}
\cite{ramakrishnan2019overcoming,talamadupula2017architectural}&\cmark & \cmark& \cmark\\
\bottomrule
\end{tabular}
\end{adjustbox}
\vspace{-0.2cm}
\end{table}
\subsection{Model Following Behavior}
While the previous section considered how the agents could use communication to address or resolve potential differences between models, in this section we will look at how agents may choose decisions that take into account multiple models. In particular, we will consider two possible classes of model following behavior. In the first case, called {\em Multi-Model Alignment} the agent chooses a behavior or decision that is not only valid in one model, but is also valid and even close to optimal in another model. 
%When there is difference between models, one way as we talked about is to communicate it to update the model. 
The second case being {\em response-seeking behavior}, wherein the agent chooses behavior based on their beliefs about the model of the other agent, so as to elicit a specific response or behavior from the other agent. (see Table \ref{tab3} as summary).\looseness=-1
\subsubsection{Multi-Model Alignment}
Current human-AI symbiosis literature have considered multiple models with which the agent could align it's decisions.

\textbf{$\mathcal{M}^*_r$ and $\mathcal{M}^R_h$ --}
An example of such alignment includes the problem studied by \cite{bansal2020optimizing}. Here the human is final decision-maker and the AI agent is making recommendation to the human. The human is expected to gradually form a mental model of when the AI agent may fail ($\mathcal{M}^R_h$), and the AI agent can potentially align its recommendation with $\mathcal{M}^R_h$ to generate decisions that are easier for the human to accept and thus optimizing for team performance. Additionally, if $\mathcal{M}^*_r$ is updated overtime and can offer better accuracy, the AI agent can align in recommendation while making a trade off between accuracy and compatibility \cite{bansal2019updates,martinez2021improving}.
Aligning AI agent's behavior to $\mathcal{M}^R_h$ has also been investigated for sequential decision-making problems under the umbrella of explicable behavior generation \cite{zhang2017plan,kulkarni2019explicable}.
In these works, the AI agent is the actor and when the two models $\mathcal{M}^*_r$ and $\mathcal{M}^R_h$ are different, the AI agent can choose to be explicable to the human by choosing plans that match human's expectation rather that directly communicating any differences. One can also try to balance between communicating the differences and aligning with the models \cite{chakraborti2017balancing,sreedharan2019expectation}. \looseness=-1

\textbf{$\mathcal{M}^*_r$ and $\mathcal{M}^H_r$ --}
One example for need for such alignment occurs when the human is the actor and the AI agent tries to come up with policies for the human. Here the AI agent needs to consider potential human limitations including sensor limitations to come up with policies that the human could execute while still maximizing its value in the true model \cite{gopalakrishnan2021synthesizing}.

\textbf{$\mathcal{M}^*_r$ and $\mathcal{M}^*_h$ --}
The AI agent and the human can both use 
their beliefs about the true task model (which may contain different information) to converge on a final decision that agrees with both of these models. We will consider this as a scenario where the decision is being aligned to both $\mathcal{M}^*_r$ and $\mathcal{M}^*_h$. One example of this is the AI-mix system that uses the crowd suggestions (i.e. information coming from $\mathcal{M}^*_h$) and criticizes generated from a partial system model (i.e. information that is part of $\mathcal{M}^*_r$)  to come up with a plan that would be better than one that could have been generated by either parties on their own. %$\mathcal{M}^*_r$ to effectively make planning and scheduling of the crowd \cite{manikonda2014ai}.

\textbf{$\mathcal{M}^*_h$ and $\mathcal{M}^R_h$ --}
This includes scenarios in which the human uses the recommendations made by the AI for their own task. As mentioned earlier, the human makes a mental model of the AI agent's error boundary that is built over time. Thus the human is effectively aligning the decision they derive using $\mathcal{M}^*_h$ with the model $\mathcal{M}^R_h$ when they accept the AI agent recommendation they believe is more likely to be true \cite{bansal2019beyond}.
\subsubsection{Response-Seeking Behavior}
% \subsubsection{\Module}
One scenario where such behavior is used is in joint workspaces. In such cases, when the AI agent comes up with its decisions, it must consider the model they ascribe to the human $\mathcal{M}^H_r$ to come up with potential coordination strategies. For example, a robot can employ human motion prediction in conjunction with a complete, time-optimal path planner to execute efficient and safe motion in a shared environment \cite{unhelkar2018human} or jointly reasoning about policy communication and acting in a collaborative manner \cite{unhelkar2020decision}. Another work that falls in this category is \cite{grover2020model}, where the AI agent uses $\mathcal{M}^H_r$ to come up with the right questions to further refine the model they ascribe to the human.
\begin{table}[t]
\centering
\caption{Summary of the various aspects that come into play in scenarios involving model following behavior}
\label{tab3}
\begin{adjustbox}{width=0.5\textwidth}
\begin{tabular}{p{17mm}|c|c|c||c}
\toprule
\multicolumn{5}{c}{Model Following Behavior}\\
\cmidrule{1-5}
\multicolumn{4}{c||}{Multi-Model Alignment} & \multicolumn{1}{c}{Response-Seeking Behavior}\\
\cmidrule{1-5}
$\mathcal{M}^*_r$ \& $\mathcal{M}^R_h$ & $\mathcal{M}^*_R$ \& $\mathcal{M}^H_r$ & $\mathcal{M}^*_r$ \& $\mathcal{M}^*_h$ & $\mathcal{M}^*_h$ \& $\mathcal{M}^R_h$ & $\mathcal{M}^H_r$\\
\cmidrule{1-5}
\cite{bansal2020optimizing,bansal2019updates,martinez2021improving,zhang2017plan,kulkarni2019explicable,chakraborti2017balancing,sreedharan2019expectation}&\cite{gopalakrishnan2021synthesizing}&\cite{manikonda2014ai} &\cite{bansal2019beyond} &\cite{unhelkar2018human,unhelkar2020decision,grover2020model} \\
\bottomrule
\end{tabular}
\end{adjustbox}
\vspace{-0.2cm}
\end{table}
\section{Gaps in Current Literature and Discussion}
\label{discussion}
In this section, we analyze the gaps in the current literature identified by our framework and how the framework connects to some of the other aspects of human-AI interaction including trust, longitudinal interaction and possible differences in inferential capabilities of humans and AI agents. Through this section we hope to provide a brief sketch on some potential future directions for research in human-AI interaction in terms of the characterization of the landscape provided here.

\subsection{Missing Interactions}
First we will consider the parts of the interactions that have received less attention in the current literature and possible opportunities that are being overlooked.
One salient observation we can make about the current landscape of work is the focus on settings that only support flow of model information from one side to be used to update the other side.
Very few current works (for example \cite{sengupta2017radar,valmeekam2020radar,grover2020radar,sengupta2018ma}) assume the possibility that both agents may have information to provide to the other agent. Arguably, this would be closer to real-life settings and an aspect that may become more apparent as we start allowing for longitudinal and iterated interaction.
In fact we would argue that the works that do support bidirectional communication are the ones that consider a form of iterative interaction. We will go over longitudinal and iterated interaction in a separate discussion section. 

As we also go over the works in the space, we also see some recurring assumptions about the models. For example, most of the works consider that the model of an agent ascribed to the other agent $\mathcal{M}^{H/R}_{r/h}$ is the same as the true task model they ascribe to the agent (i.e. it coincides with the model information they ascribe to the other agent as part of $\mathcal{M}^*_{r/h}$). Additionally, many of the works assume that the update of one agent's mental model of the other agent will also cause the agent to update what they believe the true task model of the other agent should be, for example updating $\mathcal{M}^{H}_{r}$ causes the robot to update $\mathcal{M}^{*}_{r}$. However, this assumption need not always hold and we could have scenarios that allow for the fact that some agents may believe that the other agent is misinformed about their own task model, an aspect that could also play a role in bidirectional and iterative communication between the agents.

Another aspect of under studied aspect of human-AI interaction relates to the model following behavior. In particular for multi-model alignment, there are no existing works that allow for the possibility that human would try to generate behavior that involves $\mathcal{M}^H$ and $\mathcal{M}^H_r$.
It is easy to see why people may have overlooked this problem as in such cases the human is trying to accommodate for AI agent's beliefs. However the systems might need to allow for such allowances to accurately capture human behavior in cases where human is aware of some limitation of the AI agent or know that the AI agent isn't aware of some aspects of the human model.

%\subsection{Longitudinal aspects of Human-aware AI Interaction}
\subsection{Bidirectional Communication and Longitudinal Interaction}
As mentioned earlier, we can see that most of the works in this area focus on one shot interaction or limit model updates to one direction. So, they assume the implicit or explicit model information from one model can result in the update to the other model. Many of the works that consider bidirectional updates such as \cite{sengupta2017radar,valmeekam2020radar,grover2020radar,sengupta2018ma} focus on decision-support systems where the human is the final decision-maker. 
Such cases allow for the fact that the human could potentially reject explanations provided by the AI system or could raise explanatory queries that may reveal some previously unspecified preferences.
 Another work that has bidirectional update is \cite{huang2019nonverbal} which considers the inverse of the previous setting. In this work, the robot is the final actor and it asks help from the human to update $\mathcal{M}^R$ but does so by helping human build a more accurate $\mathcal{M}^R_h$ so the human can provide more helpful information. 
 Both of these works present a limited form of interactive setting, wherein the human and AI agent converge to a final decision over a number of different iterations.
 Though one would ideally want to see systems that go beyond just these limited iterative setting and consider truly longitudinal interaction, where the human-AI collaboration may extend over a longer period of time that may cover multiple tasks. Here the results of previous interaction may influence and direct the form of future collaborations. Unfortunately, there aren't works that can support such forms of interaction. In fact one crucial aspect we will need to better model to allow for such interaction would be an effective way to model trust between the agents \cite{xu2015optimo,sengupta2019monitor}.\looseness=-1
 %Although, these works goes further by considering two direction of model update, these sequence can continue further between human and the AI agent. Thus, in longitudinal human-AI interaction, the question of when and how mental model update should be stopped give rise to the degree of trust in the human \cite{} and the degree of freedom of the AI agent model to accept inputs from the human.
\subsection{Trust in Human-AI Interaction}

In longitudinal interaction, the exact model update to be provided can also depend on the level of trust. For instance, the validity of the assumption that updates to $\mathcal{M}^R_h$, will get reflected in $\mathcal{M}^*_h$ depends on the human trust in the AI agent and their willingness to comply with the AI agent. Thus, if the human trust is low, then the difference between $\mathcal{M}^*_h$ and $\mathcal{M}^R_h$ may lead the human to think that the AI agent misunderstands the task. As a result, the AI agent's decision to provide model updates or perform model alignment should also consider the human's level of trust and it's effect on reliance and compliance from the human's end \cite{zahedi2021trust}. Furthermore, one could also have cases where if the human's trust is too high, they may accept information from AI agent and update their model when they shouldn't \cite{bansal2020does}.
All of these point to the critical role trust plays in the context of longitudinal interaction. Ideally one would want the choice for when the AI agent should provide model information or perform model following behavior and when they shouldn't, to be driven not only by the human's trust level but also based on the possible effects the AI agent's choices may have on the human trust \cite{zahedi2021trust}. All of this focuses on the human's trust and how they determine their reception to the AI agent's actions, one would also want to develop a similar mechanism at the AI agent's end to decide when they should accept information or decide to rely on the human.
\subsection{Knowledge vs. Computational Capability}
The majority of the works covered in this paper consider that the potential source of disagreement between the human and the AI agent about any aspect of the problem stems from a purely epistemic or knowledge difference between the corresponding models. However, as one considers more complex problems, an equally relevant aspect becomes the agents ability to effectively reason with the given models.
After all, humans are best captured computationally as bounded rational agent and even if they share the same model information the human and AI agent may come to differing decisions. While there are some preliminary works that tries to address this asymmetry most of them are in their infancy. 
One group of work try to use simplified model information or decisions that are easier for people to follow (examples include \cite{gopalakrishnan2021graphmin}, \cite{sreedharan2018hierarchical}). Another set of works tries to build proxy models that tries to predict how human's may react to particular model update or aligned behavior (examples include \cite{zhang2017plan}, \cite{sreedharan2019model}).

This brings us to the end of our paper. Through the introduction of the generalized human-aware AI interaction framework, we not only hope to introduce a tool to better understand the current landscape of works dealing with human-AI interaction, but also a way to identify potential limitations of current work and future research directions.
%\section{Conclusion}
%This paper provide an overview of mental-model centric landscape for human-AI symbiosis. We introduced a general version of human-aware AI interaction and saw how this new framework can allows for capturing various works done in the space of human-AI interaction. We also used this framework to highlight the behavioral pattern of works in this area and finally identified the potential gaps in the current literature and directions for future work in this area.

\noindent \paragraph{\textbf{Acknowledgments.}}This research is supported in part by ONR grants N00014- 16-1-2892, N00014-18-1- 2442, N00014-18-1-2840, N00014-9-1-2119, AFOSR grant FA9550-18-1-0067, DARPA SAIL-ON grant W911NF19-
2-0006 and a JP Morgan AI Faculty Research grant.

\bibliographystyle{plain}
\bibliography{main_extended}
\end{document}